\documentclass{ceurart}

\usepackage{listings}
\usepackage{soul}
\usepackage{url}
\usepackage[utf8]{inputenc}
\usepackage[small]{caption}
\usepackage{graphicx}
\usepackage{amsmath}
\usepackage{amsthm}
\usepackage{booktabs}
\usepackage{algorithm}
\usepackage{algorithmic}
\usepackage{enumitem}
\setlist{nosep,leftmargin=*}
\usepackage{tabularx}
\usepackage{tikz}
\usetikzlibrary{positioning, arrows.meta, shapes.geometric, fit, backgrounds}
\usepackage{xcolor}
\usepackage{siunitx}
\usepackage{multirow}
\usepackage{subcaption}
\usepackage[most]{tcolorbox}
\usepackage{etoolbox}
\usepackage{float}

\begin{document}

\copyrightyear{2026}
\copyrightclause{Copyright for this paper by its authors.
  Use permitted under Creative Commons License Attribution 4.0
  International (CC BY 4.0).}

\conference{Joint Workshop on Statistics and Knowledge Integration for Logic, Learning, Ethical Decisions, and LLMs, 18 July 2026, Lisbon}

\title{PrologMCP: A Standardized Prolog Tool Interface for LLM Agents}

\author[1]{Agnieszka Mensfelt}[%
email=agnieszka.mensfelt@rhul.ac.uk,
]
\author[1]{Adarsh Prabhakaran}[%
email=adarsh.prabhakaran@rhul.ac.uk,
]
\author[1]{Adrian Haret}[%
email=adrian.haret@rhul.ac.uk,
]
\author[1]{Vince Trencsenyi}[%
email=vince.trencsenyi@rhul.ac.uk,
]
\author[1]{Kostas Stathis}[%
email=kostas.stathis@rhul.ac.uk,
]
\address[1]{Department of Computer Science, Royal Holloway, University of London, UK}

\begin{abstract}
Frontier reasoning-tuned language models still fail on deductive tasks at depth, and the cost of improved performance through extended internal reasoning scales poorly. Symbolic delegation offers a complementary route: a language model translates the problem, while a solver performs the inference. However, current autoformalization pipelines for logic programming are typically bespoke integrations tied to particular tasks or agents. We introduce \textsc{PrologMCP}, a task-agnostic, open-source server that exposes Prolog as a stateful tool through the Model Context Protocol (MCP). Its compact tool interface, structured error reporting, and per-session isolation make the \emph{translate--run--inspect--repair} loop a reusable primitive for MCP-capable agents. We evaluate a formalizer agent enhanced with \textsc{PrologMCP} against standard and reasoning LLMs (Claude Sonnet 4.6, GPT-4.1, and o4-mini) on two subsets of \texttt{PARARULE-Plus}: a general-purpose sample and a more challenging one targeting a specific failure mode of natural-language reasoning. On the general sample, the formalizer matches or exceeds reasoning LLMs (accuracy 1.00 vs.\ 1.00 / 0.998), with the largest gains over standard models (0.762 for GPT-4.1). On the challenging subset, the formalizer remains near-perfect (1.00 / 0.99) while reasoning LLMs drop to 0.95 / 0.94. These results suggest that delegating inference to Prolog via MCP is a robust and inspectable alternative to extended natural-language reasoning.
\end{abstract}

\begin{keywords}
  Prolog \sep
  Model Context Protocol \sep
  Autoformalization \sep
  Large Language Models \sep
  Symbolic Reasoning
\end{keywords}

\maketitle

\section{Introduction}
\label{sec:introduction}

Since the dawn of artificial intelligence, reasoning has been considered a constitutive component of intelligent behaviour~\cite{mccarthy1959,kowalski1974,newell1976}. Large language models, after initially underperforming on reasoning benchmarks, have made rapid progress; however, important limitations remain. 

First, accuracy degrades on long deductive chains, even for strong reasoning-tuned models. Unlike a symbolic solver, for which multi-step rule application preserves correctness by construction, a language model has no mechanism ensuring that each generated step is a valid logical consequence of the previous one. This lack of stepwise guarantees makes long chains fragile: errors can accumulate, and successful reasoning depends on maintaining a coherent derivation. Lin et al.~\cite{lin2025zebralogic} demonstrate this collapse on grid-style logic puzzles, where frontier reasoning models suffer significant accuracy drops as problem complexity increases. Kazemi et al.~\cite{kazemi2025bbeh} likewise report that frontier reasoning models remain far from saturation on a hard reasoning benchmark. Earlier analyses of compositional tasks~\cite{dziri2023faith} and template-based variants of standard mathematics benchmarks~\cite{mirzadeh2025gsm} attribute such failures to pattern matching over training-distribution structure rather than genuine multi-step deduction, with accuracy degrading as the number of required reasoning steps grows.

Second, the cost of additional reliability scales poorly. Current reasoning-tuned models attempt to improve reliability by allocating more internal reasoning, or \textit{thinking}, tokens. However, this increases inference cost while providing no guarantee that the additional tokens correspond to valid deductive steps. On tasks whose rules are simple but whose search space is combinatorially branching, additional generation can therefore yield diminishing returns. Recent work characterising these limits~\cite{shojaee2025illusion,lin2025zebralogic} finds that increasing the thinking budget yields diminishing returns beyond a problem-complexity threshold, with reasoning models in some cases even reducing their thinking-token usage as problems become harder. By contrast, symbolic solvers---Prolog for rule-based deduction, SAT and CSP solvers for combinatorial search---routinely handle problems many orders of magnitude beyond the thresholds at which reasoning-tuned models collapse.

These solvers offer two properties that natural-language reasoning lacks: predictable compute scaling and answers that are correct by construction. This motivates an established alternative---the tool-use paradigm~\cite{schick2023toolformer,gao2023pal}---in which, rather than perform every reasoning step internally, a model delegates well-specified tasks to external tools. For instance, arithmetic and general computation can be handled by code interpreters; factual lookup by retrieval systems; and reasoning problems by symbolic solvers. This latter direction is closely related to \emph{autoformalization}~\cite{wu_autoformalization_2022,mensfelt2026towards}, which has explored several classes of formal systems and solvers. Logic programming is an especially natural fit for many deductive-reasoning tasks, since it combines declarative representation with automated inference. Yet, compared with code execution and retrieval, logic programming remains comparatively under-served as a standardized tool for language-model agents.

A growing line of work has explored combinations of LLMs and Prolog. \cite{borazjanizadeh2024reliable} use an off-the-shelf LLM to generate Prolog programs and re-try on execution failure; \cite{mensfelt2024generative} autoformalize game-theoretic scenarios into executable Prolog with an iterative solver-feedback correction loop---a close precedent for the \textit{translate--run--inspect--repair} pattern we propose here. \cite{yang2024arithmetic} fine-tune LLMs to generate Prolog for arithmetic word problems, while \cite{tan2024thought} use a Prolog engine to produce verified reasoning trajectories that an LLM imitates as structured chain-of-thought. Closely related, \cite{mellgren2025training} fine-tune a small model with GRPO to use Prolog as a callable tool with internal repair, motivating the primitive that \textsc{PrologMCP} exposes as a standardized interface. Other recent work~\cite{zunjare2026neuroprolog,he2025protoreasoning} incorporates structured Prolog representations during training.

However, existing systems are largely bespoke: each defines its own interface, execution model, and error-handling strategy, limiting portability across tasks and agent frameworks. Several open-source Prolog MCP servers also exist (e.g., \cite{rybinski,umuro,rikarazome,vpursuit}), exposing query execution and, in some cases, persistent knowledge bases. To our knowledge, however, none provides structured diagnosis of silent failures
nor reports an extensive evaluation of a formalizing agent on a reasoning benchmark.
This paper presents \textsc{PrologMCP}, a Model Context Protocol (MCP) server that exposes Prolog as a stateful tool supporting a reusable \emph{translate--run--inspect--repair} loop. Our contributions are:

\begin{enumerate}[label=\Roman*.]

\item \textbf{A standardized Prolog tool interface} implemented as an MCP server, providing—beyond query execution—structured proof and failure trees, predicate inspection, hot clause replacement, and session-level isolation that together support agent-driven debugging (Section~\ref{sec:architecture}).

\item \textbf{A formalizing LLM agent}, translating natural-language problems into Prolog, executing queries using the server, and iteratively refining its representations using structured feedback (Section~\ref{sec:evaluation}).

\item \textbf{An empirical evaluation of the formalizer agent} on \texttt{PARARULE-Plus}~\cite{bao2022multi}, comparing it against standard LLM and reasoning-model inference (Section~\ref{sec:evaluation}).

\end{enumerate}

\section{Background}

\paragraph{Logic Programs in Prolog.}
Logic programs, most prominently implemented in Prolog~\cite{korner2022fifty,wielemaker2012swi}, provide a declarative representation based on Horn clauses: universally quantified implications of the form ``if $b_1$, \dots, $b_n$, then $h$'', which map naturally onto facts (clauses with no body), rules (clauses with one or more body literals), and queries (goals to be proved). Reasoning combines \textit{unification} of terms, \textit{SLD-resolution}~\cite{kowalski1971linear} to reduce a query to subgoals against matching clause heads, and goal-directed search with backtracking. Prolog extends SLD with SLDNF, i.e., negation-as-failure~\cite{clark1978negation}, whereby a negated goal is treated as satisfied when the system cannot prove the corresponding positive goal.

Horn clauses align closely with the surface form of natural-language rules (``if X is kind then X is nice'' becomes \texttt{nice(X) :- kind(X).}), and Prolog's goal-directed inference, support for negation-as-failure, and persistent clause state across calls make it a natural target for autoformalization in deductive-reasoning tasks. Competing approaches such as SAT, SMT, or CP require translation into satisfiability formulas, background theories, or explicit variable domains, and SAT and CP solvers are typically stateless per invocation, which fits multi-turn agent reasoning over a growing knowledge base less well.

\paragraph{Model Context Protocol.}
The Model Context Protocol~\cite{anthropic2024mcp,Hou2026e} provides a standard interface for exposing tools to LLM agents, with conventions for tool discovery, typed argument schemas, and structured error reporting. A tool exposed via MCP can be used by any MCP-aware agent without model-specific glue code. MCP also accommodates stateful tools such as Prolog: across a sequence of calls, an agent can assert clauses, inspect program state, issue queries, receive structured diagnostics, and refine the formalization. The closest precedent is the MCP-Solver discussed in~\cite{szeider2025bridging}, which exposes constraint, SAT, SMT and answer sets backends (MiniZinc, PySAT, Z3 and clingo) but not Prolog-style SLDNF resolution with unification and backtracking over a database of clauses.

\paragraph{Autoformalization.}
\emph{Autoformalization} is the automatic translation of an informal-language expression into a semantically corresponding expression in a formal language. The term originates in formal mathematics~\cite{wu_autoformalization_2022}, and has since broadened to cover pipelines in which a model emits a formal artefact---a logic program, constraint model, SMT formula, or planning specification---passed to a sound external reasoning procedure. A recent unifying account~\cite{mensfelt2026towards} parameterises autoformalization by an informal language $L_i$ and a formalizable subset thereof, a formal reasoning language $L_f$ with precise semantics, and a semantic equivalence criterion $E$ specifying which aspects of meaning must be preserved. Because $E$ is itself informal and not directly checkable, practical pipelines introduce a computable \emph{validation criterion} $V$ that approximates $E$, for example by executing the formal artefact with a solver and comparing its output against an expected result.

\begin{figure}[t]
  \centering
  \includegraphics[width=0.75\linewidth]{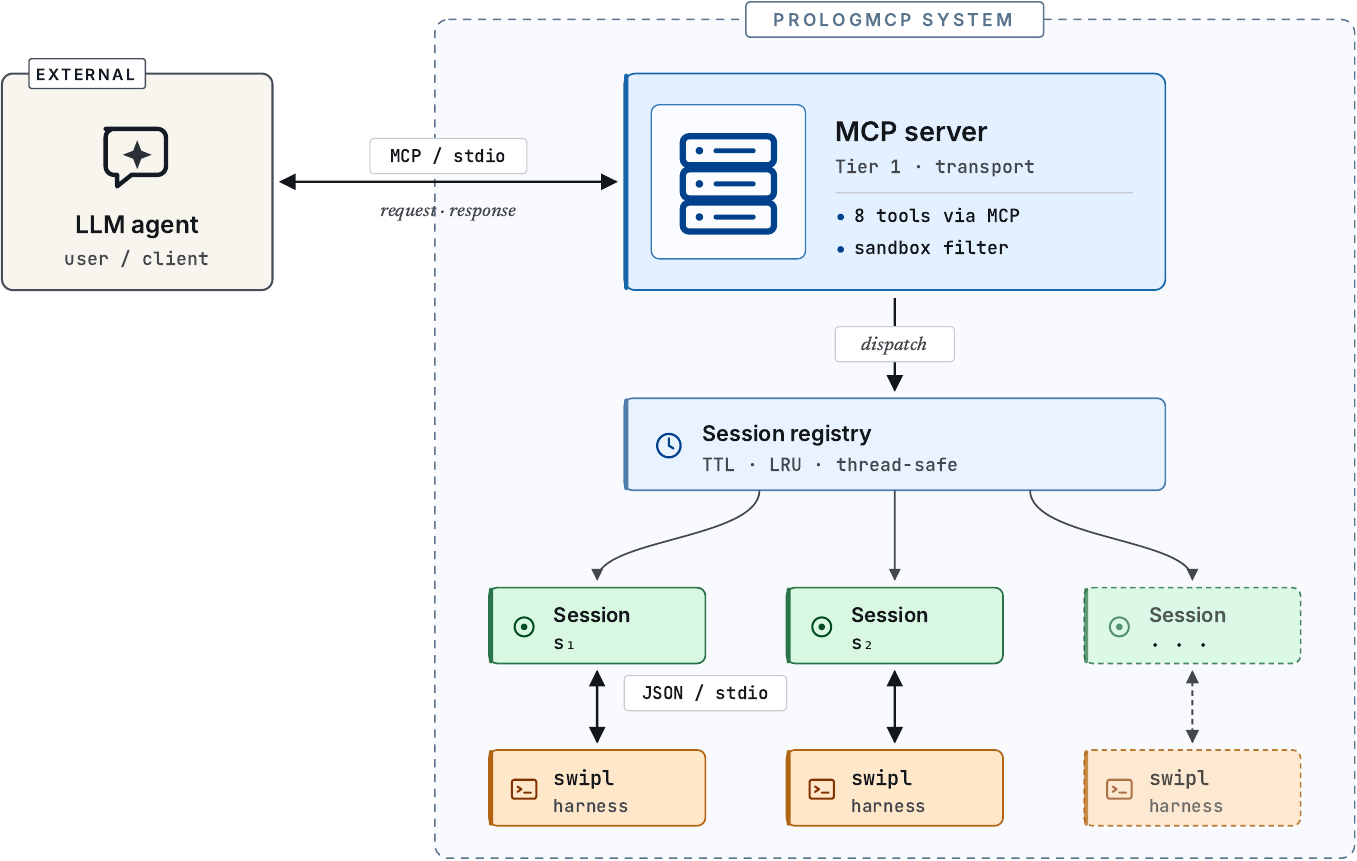}
  \caption{PrologMCP architecture.}
  \label{fig:architecture}
\end{figure}

\section{Architecture}
\label{sec:architecture}

\textsc{PrologMCP} is an MCP server that exposes Prolog to LLM agents as a stateful tool. Three properties drive the design: ($i$) \emph{statefulness}---clauses persist across calls, so the agent builds a knowledge base incrementally instead of resubmitting it on every goal; ($ii$) \emph{structured error reporting}---compile- and runtime diagnostics are returned as typed JSON; ($iii$) \emph{isolation}---each session runs in a dedicated interpreter subprocess with bounded resources. Currently, the server supports only SWI-Prolog. Tool signatures and JSON schemas are given in Appendix~\ref{app:tools}. The source code is available on GitHub (\url{https://github.com/dicelab-rhul/PrologMCP}).

\subsection{Components}
\label{sec:arch:components}

The system has three tiers (Figure~\ref{fig:architecture}), forming a single path from an agent's tool call down to a Prolog subprocess and back:

\begin{enumerate}

  \item \textbf{A Python MCP server} that presents the tool interface of Section~\ref{sec:arch:tools} to the agent: it declares each tool's input schema, validates and dispatches incoming calls to the appropriate session, and maps harness outcomes---including failures---onto MCP's structured tool-result format. The server holds no Prolog state of its own; it is a thin protocol adapter over the session registry.
  \item  \textbf{A session registry} that owns a pool of Prolog sessions, indexed by a \texttt{load\_id} returned on session creation. The registry is thread-safe, caps concurrent sessions at a configurable limit, evicts LRU (least-recently-used) sessions on overflow, and reaps sessions idle beyond a TTL (time-to-live) from a background thread.
  \item  \textbf{A Prolog harness} (\texttt{harness.pl}), loaded into each session's \texttt{swipl} subprocess, that reads newline-delimited JSON commands on \texttt{stdin}, dispatches them to handlers, and writes newline-delimited JSON responses on \texttt{stdout}. Stderr is reserved for uncaptured diagnostics.
\end{enumerate}

\subsection{Session Semantics}
\label{sec:arch:sessions}

A session is created by \texttt{consult\_text(source, dialect)}: the registry allocates a fresh \texttt{load\_id}, spawns a \texttt{swipl} subprocess with the harness, writes \texttt{source} to a file in the session's temporary directory, and issues a \texttt{consult} command. If \texttt{source} lacks a module declaration, the registry prepends \texttt{:-~module(m\_\textit{\textlangle load\_id\textrangle}, []).} so that subsequently defined predicates live in a per-session module disjoint from the harness and from other sessions. Module-qualified goal evaluation (\texttt{Module:Goal}) is used throughout, so predicate resolution is unambiguous regardless of the module in which \texttt{source} was written.

Syntax errors, singleton warnings, and similar diagnostics are captured by a \texttt{message\_hook/3} installed in the harness, so a consult that raises errors still loads the clauses it can rather than aborting. Captured diagnostics are returned as a \texttt{messages} array, each entry carrying a severity, a kind drawn from a fixed vocabulary, a message string, and, when the diagnostic is attributable to a specific clause, the source line that triggered it. Because the successfully loaded predicates remain callable after a failed consult, the agent can repair the source with \texttt{replace\_predicate}; consultation is thus \emph{non-fatal}.

The subprocess is held open for the session's lifetime; sessions close explicitly via \texttt{close\_session}, implicitly on TTL expiry, or by LRU eviction under load.

\subsection{Tool Interface}
\label{sec:arch:tools}

\begin{table}[tbp]
  \centering
  \small
  \begin{tabularx}{\linewidth}{@{}l X@{}}
    \toprule
    \textbf{Tool} & \textbf{Behaviour} \\
    \midrule
    \texttt{consult\_text}
      & Open a session from a source string, returning its \texttt{load\_id}, an optional inventory of defined predicates, and any diagnostics raised during consultation. \\
    \addlinespace
    \texttt{run\_goal}
      & Evaluate a goal string with \texttt{findnsols/4} wrapped in \texttt{call\_with\_depth\_limit/3}, reporting typed bindings, the number of solutions, and whether the depth bound was hit. \\
    \addlinespace
    \texttt{inspect\_predicate}
      & Report a predicate's clauses, arity, type (\texttt{static\,|\,dynamic\,|\,foreign}), export status, and clause count; if no exact match exists, suggest predicates with similar names. \\
    \addlinespace
    \texttt{get\_source}
      & Rebuild the module's source by traversing \texttt{predicate\_property/2} and \texttt{clause/2}. \\
    \addlinespace
    \texttt{replace\_predicate}
      & Retract a predicate with \texttt{abolish/1}, then reload a single-predicate patch into the session module via \texttt{load\_files/2}. \\
    \addlinespace
    \texttt{list\_messages}
      & Retrieve the diagnostics gathered over the current session. \\
    \addlinespace
    \texttt{run\_tests}
      & Consult an optional \texttt{plunit} file and execute every registered unit, returning pass, fail, and blocked tallies alongside per-test results. \\
    \addlinespace
    \texttt{trace\_goal}
      & Run a depth-bounded meta-interpreter over a goal, yielding a proof tree and its node count. \\
    \midrule
    \texttt{close\_session}
      & Tear down the subprocess and remove its temporary directory. \\
    \bottomrule
  \end{tabularx}
  \caption{Tools exposed over MCP. All arguments and return values are
    JSON; every error is reported as
    \texttt{\{ok:\,false,\,error:\,\{kind,\,message,\,prolog\_term\}\}}
    without terminating the session.}
  \label{tab:tools}
\end{table}

The server exposes eight operational tools plus one administrative tool (Table~\ref{tab:tools}).
Each tool corresponds to one primitive the harness supports directly, and the Python layer does no orchestration of its own beyond dispatch and error mapping.

We bound goal evaluation in two ways: by the number of solutions returned per call and by inference depth. A goal that exhausts the depth limit returns with \texttt{depth\_exceeded:\,true} and any solutions found so far, rather than raising an exception. Both bounds are surfaced in the response, so the agent can retry with a larger limit when a partial result is inconclusive. These bounds are surfaced in the response so that the agent can retry with a larger limit if the partial result is inconclusive.

\section{Evaluation}
\label{sec:evaluation}

\subsection{Agent implementation}
\label{sec:evaluation:client}

\begin{figure}[tbp]
    \centering
    \includegraphics[width=0.75\linewidth]{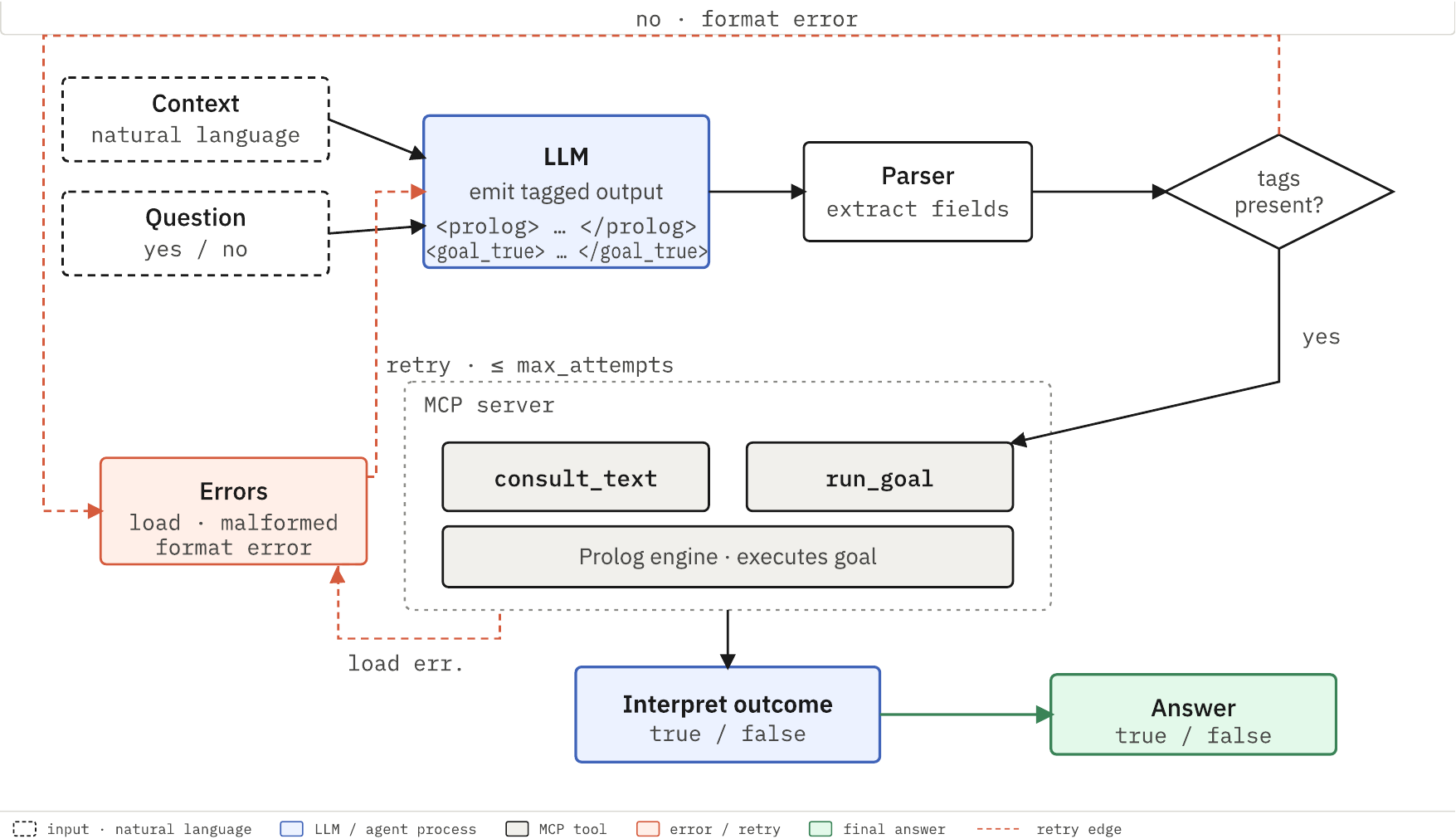}
    \caption{Architecture of the Prolog-augmented Formalizer agent.}
    \label{fig:formalizer}
\end{figure}

We evaluate three agents that differ in how they produce a final answer: a \textbf{Standard} baseline, a \textbf{Reasoning} baseline, and a Prolog-augmented \textbf{Formalizer}. All agents receive the same natural-language context and yes/no question. \textbf{Standard} is a single-call baseline. The model receives the context and question, prefaced by a system prompt that frames the task as logical reasoning under the \texttt{PARARULE-Plus} negation semantics. It is instructed to answer with exactly one word, \texttt{true} or \texttt{false}, without elicited intermediate reasoning. \textbf{Reasoning} uses the same task formulation as the Standard baseline, but runs it through each provider's reasoning-augmented inference path. For Anthropic models, this corresponds to Extended Thinking mode; for OpenAI models, we use a dedicated reasoning model in place of the standard GPT model.

\textbf{Formalizer} delegates inference to \textsc{PrologMCP}. Given the natural-language context and question, the model generates a Prolog program together with a query goal. The agent loads the generated program through the MCP \texttt{consult\_text} tool, executes the query through \texttt{run\_goal}, and interprets the result (Figure~\ref{fig:formalizer}). To make this interaction robust, the model is instructed to emit the program and goal inside \texttt{<prolog>}\ldots\texttt{</prolog>} and \texttt{<goal\_true>}\ldots\texttt{</goal\_true>} tags. A parser extracts these fields; missing or malformed fields are treated as format errors. If parsing fails, if the Prolog program cannot be loaded, or if execution reports an error, the agent returns the structured error message to the model and retries, up to \texttt{max\_attempts} attempts.

The source code for the evaluation framework and experimental logs are available on GitHub (\url{https://github.com/dicelab-rhul/MCPFormalizer}). The prompts used in the evaluation are provided in Appendix~\ref{app:prompts}: the Formalizer prompt appears in Appendix~\ref{app:prompts:formalize}, and the Standard prompt appears in Appendix~\ref{app:prompts:standard}.

\subsection{Dataset}

The \texttt{PARARULE-Plus} dataset~\cite{bao2022multi} is a large-scale synthetic dataset for multi-step deductive reasoning over natural language, extending the original PARARULE~\cite{Clark2020p} with deeper inference chains. Each instance is a set of natural-language facts and rules plus a Boolean query (examples in Appendix~\ref{app:examples}). \texttt{PARARULE-Plus} covers reasoning depths 2--5 with $\approx$\num{100000} samples each, generated over simple people- and animal-attribute domains.

The dataset uses a specific, non-standard interpretation of negation. A body-level \texttt{\textbackslash+\,P(x)} succeeds iff \texttt{P(x)} is not in the \emph{initial} fact base, regardless of whether it is derivable by rules---resembling negation in semi-positive Datalog~\cite{abiteboul1995foundations}, where negation is restricted to the extensional database. Queries, by contrast, use the standard closed-world reading over both extensional and intensional facts: a negated query \texttt{\textbackslash+\,P(x)} is true iff \texttt{P(x)} is neither stated nor derivable. Both standard SLDNF and stratified negation~\cite{AptBlairWalker1988} would instead apply the closure uniformly at both levels. This non-standard reading is required to reproduce the ground-truth labels and is passed on to the LLMs in the prompts (Appendix~\ref{app:prompts}). This reading is passed to the LLMs by instructing the formalizer agent to record each base fact additionally as \texttt{initially(pred, entity)} and to render every body-level negation as \texttt{\textbackslash+\,initially(pred, X)} (keeping \texttt{initially/2} out of the goal), and by telling the standard and reasoning LLMs to check body-level negations against initial facts only but negated questions against the full closure.

\subsection{Experimental setup}

\begin{table}[tbp]
\centering
\begin{subtable}[t]{0.5\linewidth}
\centering
\small
\begin{tabular}{llc}
\toprule
System & Model & Reasoning budget \\
\midrule
Standard & \texttt{sonnet-4-6} & --- \\
Reasoning & \texttt{sonnet-4-6} & 10{,}000 tokens \\
Formalizer & \texttt{sonnet-4-6} & --- \\
\midrule
Standard & \texttt{gpt-4.1} & --- \\
Reasoning & \texttt{o4-mini} & medium effort \\
Formalizer & \texttt{gpt-4.1} & --- \\
\bottomrule
\end{tabular}
\caption{Per-system model and reasoning-budget settings.}
\label{tab:eval-params-system}
\end{subtable}%
\hfill
\begin{subtable}[t]{0.5\linewidth}
\centering
\small
\begin{tabular}{ll}
\toprule
Parameter & Value \\
\midrule
Random seed & 42 \\
Max formalization attempts & 5 \\
Max tokens (Standard, Formalizer) & 4{,}096 \\
Max tokens (Reasoning) & 11{,}024 \\
Temperature (Standard, Formalizer) & 0 \\
Temperature (Reasoning) & n/a \\
\bottomrule
\end{tabular}
\caption{Global parameters held constant across all conditions.}
\label{tab:eval-params-global}
\end{subtable}
\caption{Experimental configuration.}
\label{tab:eval-params}
\end{table}

We evaluate each agent on two frontier models, Anthropic's Claude Sonnet 4.6 and OpenAI's GPT-4.1, with the Reasoning condition substituting o4-mini in place of GPT-4.1 on the OpenAI side. Per-system parameters are summarised in Table~\ref{tab:eval-params-system} and global parameters in Table~\ref{tab:eval-params-global}. Temperature is fixed at zero for the Standard and Formalizer conditions to make outputs deterministic, but cannot be set uniformly across the evaluation: Anthropic's Extended Thinking and OpenAI's o4-mini both disable temperature control by API constraint, exposing only their respective reasoning-budget parameters in its place. The two Reasoning conditions are therefore compared at the qualitative setting recommended by each provider, namely a \num{10000}-token thinking budget for Extended Thinking and medium reasoning effort for o4-mini.

The MCP server was started fresh per process via the \texttt{prolog-mcp} CLI entry point, and each instance received its own isolated Prolog session, created by \texttt{consult\_text} and closed on completion. Accuracy is reported as the fraction of instances where the predicted answer matches the ground-truth label, with instances returning \texttt{error} or \texttt{unknown} counted as incorrect; no instances returned \texttt{error} in any run.

We evaluate \textsc{PrologMCP} along two complementary axes. Section~\ref{sec:depth-stratified} reports results on a depth-stratified sample of \texttt{PARARULE-Plus}, reflecting the natural distribution of the benchmark across reasoning depths and rule types. Section~\ref{sec:not-sldnf} then reports results on a targeted subset that isolates instances where the dataset's negation semantics provably diverges from standard SLDNF evaluation. Together, these evaluations test both the average-case reliability of symbolic delegation and its robustness under adversarial semantic conditions.

\section{Results}
\label{sec:results}

\subsection{Depth-stratified evaluation}
\label{sec:depth-stratified}

\paragraph{Dataset} We sampled 400 instances from \texttt{PARARULE-Plus}, stratified by reasoning depth, rule type, and domain. Although \texttt{PARARULE-Plus} differs from SLDNF in its treatment of negation, the sampled subset contains substantial overlap between the two semantics because many body-level negations occur in cases where the distinction is not observable. In 45\% of negated predicates, the negated predicate appears only as a base fact and is never derived by any rule. For such predicates, checking absence from the initial facts is equivalent to checking failure of proof, since no rule can derive the predicate. In the remaining 55\%, the negated predicate does have derivation rules, so the two semantics can in principle disagree. However, for many specific entities, those rules do not derive the predicate being negated. The proof-theoretic difference is therefore present but not activated by the particular query and knowledge base. Disagreement arises only when the negated predicate is both derivable by rules and actually derived for the entity under consideration; this occurs in 6 instances. Non-termination would also arise when proving the negated predicate requires exploring recursive rule cycles under SLDNF; this would occur in 15 instances.

\paragraph{Main results.}

\begin{table}[tbp]
\small
  \centering
  \begin{tabular}{lrrrrrr}
    \toprule
    System & Accuracy & Input  & Output & Total  &  Sec.\\
    \midrule
    \multicolumn{6}{c}{Claude} \\
    \midrule
    Standard LLM& 0.988 & \textbf{688} & \textbf{160} & \textbf{849} & \textbf{3.57} \\
    Reasoning & \textbf{1.000} & 717 & 240 & 957 & 4.64 \\
    Formalizer (Prolog) & \textbf{1.000} & 1003 & 423 & 1426 & 5.11 \\
    \midrule
    \multicolumn{6}{c}{GPT/o4-mini} \\
    \midrule
    Standard LLM & 0.762 & 653 & \textbf{2} & \textbf{655} & \textbf{0.98} \\
    Reasoning & 0.998 & \textbf{652} & 896 & 1548 & 11.27 \\
    Formalizer (Prolog) & \textbf{1.000} & 881 & 308 & 1189 & 4.23 \\
    \bottomrule
  \end{tabular}
  \caption{Overall results on \texttt{PARARULE-Plus} (400 instances).}
  \label{tab:pararule_main}
\end{table}

Table~\ref{tab:pararule_main} reports the main results for the three systems described in Section~\ref{sec:evaluation:client}: the Standard baseline, which answers without an explicit reasoning configuration; the Reasoning condition, which uses a reasoning-enabled inference path or dedicated reasoning model; and the Formalizer, which translates the instance into Prolog and delegates inference to the MCP server.

Overall, both reasoning-augmented approaches improve accuracy over standard LLMs, with the effect most pronounced for the OpenAI models. For Claude Sonnet 4.6, the Standard baseline is already strong, achieving 0.988 accuracy while using the fewest tokens and lowest runtime. Both the Reasoning condition and the Formalizer reach perfect accuracy, but at higher computational cost. One caveat is that, despite the prompt instruction to answer with a single word, Claude Sonnet 4.6 in the Standard condition sometimes produces explanatory reasoning traces. This increases its output-token usage and makes the condition less sharply separated from the explicit reasoning setting.

For the OpenAI models, GPT-4.1 follows the one-word response instruction more closely in the Standard condition, but its accuracy is much lower, at 0.762. Replacing GPT-4.1 with the reasoning model o4-mini raises accuracy to 0.998, while the GPT-4.1 Formalizer reaches perfect accuracy. The Formalizer is particularly effective in this setting: it achieves the best accuracy with lower total token usage and substantially lower runtime than the reasoning-model condition. These results suggest that symbolic delegation can provide a favourable reliability--cost trade-off when the base model is less reliable, while direct natural-language inference may remain more efficient when the base model is already highly accurate.

\paragraph{Accuracy by depth.}

\begin{table}[tbp]
\small
\centering
\begin{subtable}[t]{0.55\linewidth}
\centering
\small
\begin{tabular}{lrrrr}
\toprule
System & $d=2$ & $d=3$ & $d=4$ & $d=5$ \\
\midrule
\multicolumn{5}{c}{Claude} \\
\midrule
Standard LLM & 1.000 & 1.000 & 0.970 & 0.980 \\
Reasoning & 1.000 & 1.000 & 1.000 & 1.000 \\
Formalizer (Prolog) & 1.000 & 1.000 & 1.000 & 1.000 \\
\midrule
\multicolumn{5}{c}{GPT/o4-mini} \\
\midrule
Standard LLM & 0.740 & 0.770 & 0.770 & 0.770 \\
Reasoning & 1.000 & 1.000 & 1.000 & 0.990 \\
Formalizer (Prolog) & 1.000 & 1.000 & 1.000 & 1.000 \\
\bottomrule
\end{tabular}
\caption{Accuracy by reasoning depth.}
\label{tab:pararule_depth}
\end{subtable}%
\hfill
\begin{subtable}[t]{0.43\linewidth}
\centering
\small
\begin{tabular}{lrr}
\toprule
System & Negation & Standard \\
\midrule
\multicolumn{3}{c}{Claude} \\
\midrule
Standard LLM & 0.975 & 1.000 \\
Reasoning & 1.000 & 1.000 \\
Formalizer (Prolog) & 1.000 & 1.000 \\
\midrule
\multicolumn{3}{c}{GPT/o4-mini} \\
\midrule
Standard LLM & 0.670 & 0.855 \\
Reasoning & 0.995 & 1.000 \\
Formalizer (Prolog) & 1.000 & 1.000 \\
\bottomrule
\end{tabular}
\caption{Accuracy by rule type ($n=200$ per cell).}
\label{tab:pararule_rule}
\end{subtable}
\caption{Per-system accuracy broken down by reasoning depth and by rule type.}
\label{tab:pararule_breakdown}
\end{table}

The depth-wise results in Table~\ref{tab:pararule_depth} show that standard models become less reliable as reasoning depth increases, whereas both reasoning-augmented approaches remain robust. For Claude Sonnet 4.6, the Standard baseline is perfect at depths 2 and 3, but drops slightly at depths 4 and 5. By contrast, both the Reasoning condition and the Prolog Formalizer achieve perfect accuracy at every depth.

The effect is stronger for the OpenAI models. GPT-4.1 standard prompting remains below 0.80 at every depth, while o4-mini Reasoning reaches perfect accuracy through depth 4 and drops only slightly at depth 5. The GPT-4.1 Formalizer achieves perfect accuracy across all depths, suggesting that explicit symbolic formalization provides the most stable performance as reasoning complexity increases.

One notable exception to a simple depth-scaling pattern is GPT-4.1 in the Standard condition: its lowest accuracy occurs at depth 1 rather than at the deepest levels. This suggests that the Standard model's errors are not driven solely by reasoning-chain length. Instead, they may reflect reliance on shallow pattern matching, rather than systematic multi-step deduction.

\paragraph{Accuracy by rule type.}

\begin{figure}[tbp]
    \centering
    \hspace*{\fill}
    \begin{subfigure}{0.35\linewidth}
        \centering
        \includegraphics[width=\linewidth]{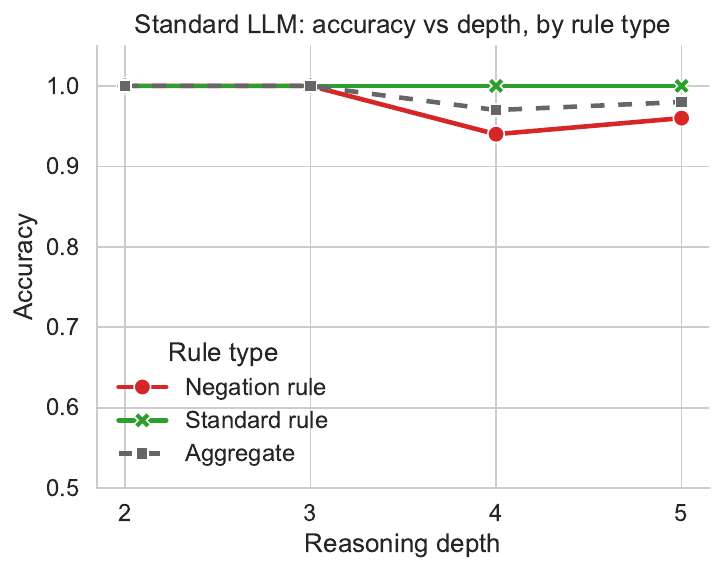}
        \caption{Claude}
        \label{fig:depth_ruletype_claude}
    \end{subfigure}
    \hfill
    \begin{subfigure}{0.35\linewidth}
        \centering
        \includegraphics[width=\linewidth]{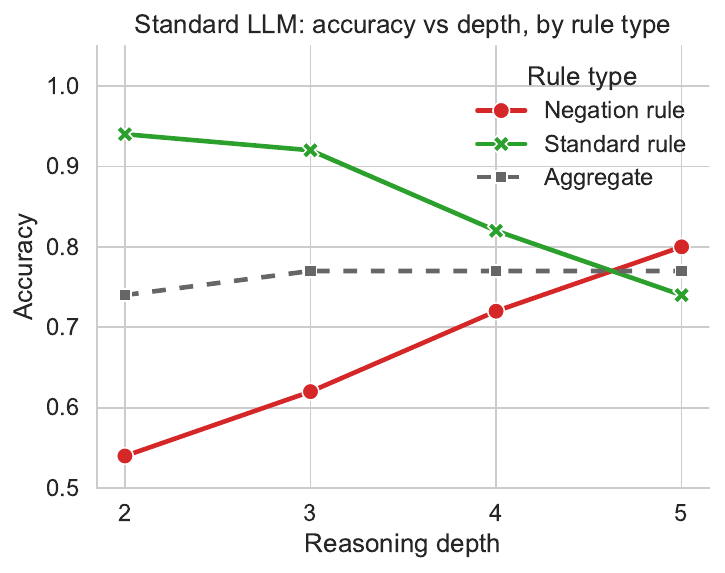}
        \caption{GPT/o4-mini}
        \label{fig:depth_ruletype_gpt}
    \end{subfigure}
    \hspace*{\fill}
    \caption{Standard-mode accuracy by reasoning depth and rule type.}
    \label{fig:depth_ruletype}
\end{figure}

Table~\ref{tab:pararule_rule} reports accuracy by rule type, and Figure~\ref{fig:depth_ruletype} reports accuracy by rule type across reasoning depths. For Claude Sonnet 4.6 in both the Standard and Reasoning conditions, and for o4-mini in the Reasoning condition, all errors occur in instances that contain negation in the rules. For Claude Sonnet 4.6 in the Standard condition, these failures are concentrated at depths 4 and 5 (Figure~\ref{fig:depth_ruletype_claude}). This pattern suggests that the non-standard negation semantics of \texttt{PARARULE-Plus} may pose a challenge even for strong reasoning models, because it requires them to suppress more familiar interpretations of negation.

GPT-4.1 in the Standard condition is the only setting in which errors also affect instances without rule-level negation. Figure~\ref{fig:depth_ruletype_gpt} shows accuracy by depth for both rule types. For GPT-4.1 Standard, accuracy on instances with negation increases with reasoning depth, while accuracy on instances without negation decreases. This non-monotonic pattern is consistent with the hypothesis that GPT-4.1 Standard relies partly on shallow regularities in the input distribution, rather than performing systematic deductive reasoning.

\paragraph{Token usage by depth.}

\begin{figure}[tbp]
    \centering
    \hspace*{\fill}
    \begin{subfigure}{0.35\linewidth}
        \centering
        \includegraphics[width=\linewidth]{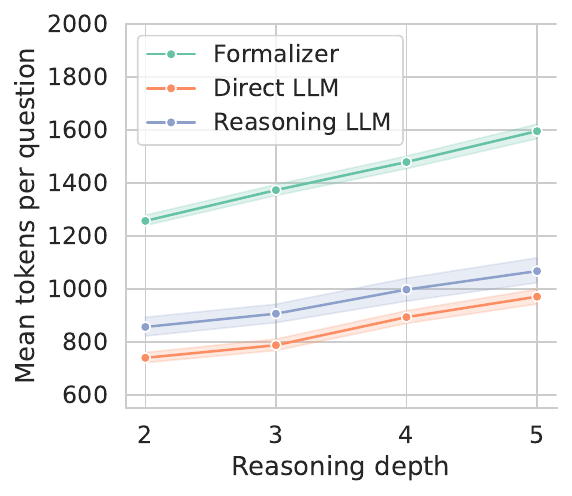}
        \caption{Claude}
        \label{fig:tokens_claude}
    \end{subfigure}
    \hfill
    \begin{subfigure}{0.35\linewidth}
        \centering
        \includegraphics[width=\linewidth]{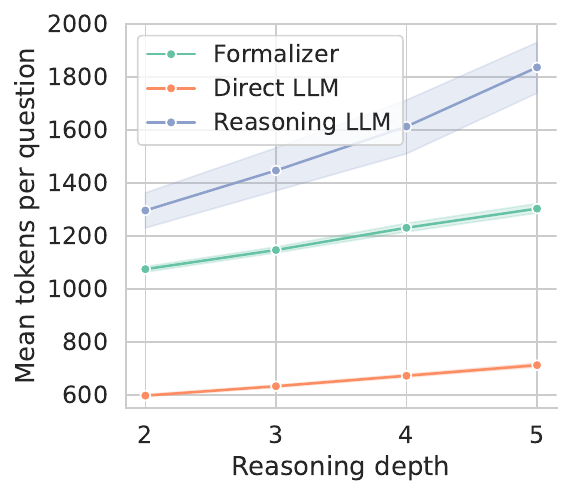}
        \caption{GPT/o4-mini}
        \label{fig:tokens_gpt}
    \end{subfigure}
    \hspace*{\fill}
    \caption{Token usage by reasoning depth for Claude and GPT models.}
    \label{fig:tokens}
\end{figure}

Figure~\ref{fig:tokens} shows token usage by reasoning depth. Token consumption increases with depth for all systems, but the magnitude of this growth differs substantially. For Claude Sonnet 4.6, total token usage rises moderately across all methods: Standard increases from 741 tokens at $d=2$ to 971 at $d=5$, Reasoning from 857 to 1068, and the Formalizer from 1257 to 1595. For the OpenAI systems, the increase is steepest in the Reasoning condition, where o4-mini grows from 1296 tokens at $d=2$ to 1835 at $d=5$. By contrast, the GPT-4.1 Formalizer increases more moderately, from 1075 to 1303 tokens, while GPT-4.1 Standard remains relatively flat, ranging from 598 to 713 tokens across depths.

Overall, reasoning-based approaches, especially o4-mini Reasoning, incur a substantial token overhead as depth increases. The Formalizer uses more tokens than Standard models but its token usage grows predictably with context length and shows the lowest variation across questions. This suggests that symbolic delegation provides a more controlled token-scaling profile than extended natural-language reasoning, while still achieving higher reliability than Standard models.

\paragraph{Runtime by depth.}

\begin{figure}[tbp]
    \centering
    \hspace*{\fill}
    \begin{subfigure}{0.35\linewidth}
        \centering
        \includegraphics[width=\linewidth]{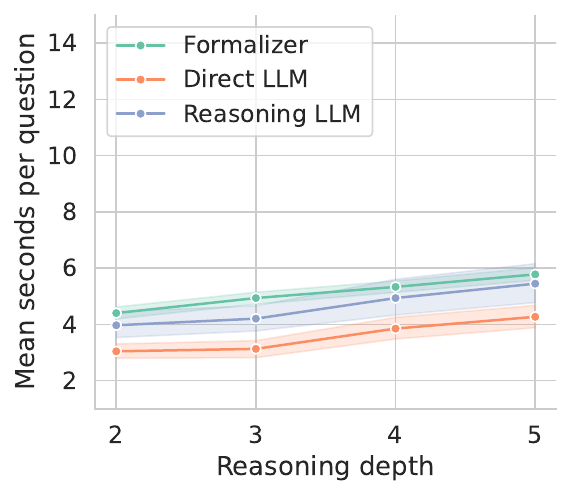}
        \caption{Claude}
        \label{fig:time_claude}
    \end{subfigure}
    \hfill
    \begin{subfigure}{0.35\linewidth}
        \centering
        \includegraphics[width=\linewidth]{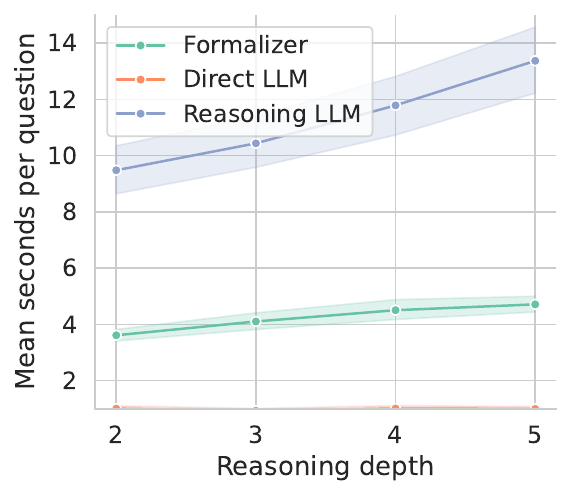}
        \caption{GPT}
        \label{fig:time_gpt}
    \end{subfigure}
    \hspace*{\fill}
    \caption{Time elapsed by reasoning depth for Claude and GPT models.}
    \label{fig:time}
\end{figure}

Execution time follows a similar trend to token usage, increasing with reasoning depth for all systems, but with substantial differences in scaling (Figure~\ref{fig:time}). For Claude Sonnet 4.6, runtime grows gradually across all methods: Standard increases from 3.04 seconds at $d=2$ to 4.27 seconds at $d=5$, Reasoning from 3.97 to 5.45 seconds, and the Formalizer from 4.40 to 5.78 seconds. At depths 3--5, the mean runtime of the Reasoning condition approaches that of the Formalizer, suggesting that the additional overhead of symbolic delegation becomes comparatively smaller as reasoning depth increases.

For the OpenAI systems, the contrast is sharper. The o4-mini Reasoning condition rises from 9.48 seconds at $d=2$ to 13.37 seconds at $d=5$, reflecting the latency cost of extended reasoning. By contrast, the GPT-4.1 Formalizer increases only from 3.61 to 4.71 seconds, while GPT-4.1 Standard remains nearly constant at around one second across depths. These results indicate that deeper reasoning consistently increases computational cost, but that symbolic delegation has substantially better runtime scaling than the reasoning-model condition, particularly when the non-tool baseline model is less reliable.

\begin{table}[tbp]
  \centering

  \setlength{\tabcolsep}{5pt}
  \renewcommand{\arraystretch}{1.15}

  \makebox[0.7\linewidth][c]{%
    \footnotesize
    \begin{tabular}{lcc}
      \toprule
      System & Correct & Incorrect \\
      \midrule

      \multicolumn{3}{c}{Claude} \\
      \midrule
      Standard          & $843 \pm 143$ $(n=395)$  & $1281 \pm 142$ $(n=5)$ \\
      Reasoning    & $957 \pm 222$ $(n=400)$  & -- $(n=0)$ \\
      Formalizer   & $1426 \pm 173$ $(n=400)$ & -- $(n=0)$ \\
      \midrule
      \multicolumn{3}{c}{GPT/o4-mini} \\
      \midrule
      Standard        & $656 \pm 53$ $(n=305)$   & $649 \pm 48$ $(n=95)$ \\
      Reasoning    & $1547 \pm 506$ $(n=399)$ & $1880$ $(n=1)$ \\
      Formalizer & $1189 \pm 113$ $(n=400)$ & -- $(n=0)$ \\
      \bottomrule
    \end{tabular}%
  }
  \caption{Token usage (mean $\pm$ std) by system and evaluation outcome.}
  \label{tab:token_stats}
\end{table}

\paragraph{Token usage by correctness.}
Table~\ref{tab:token_stats} reports token usage separately for correct and incorrect predictions. Except for GPT-4.1 in the Standard condition, incorrect answers use more tokens on average than correct answers. This pattern is consistent with prior observations that reasoning models often generate more tokens on harder instances~\cite{lin2025zebralogic}. However, the number of failures in the reasoning-augmented conditions is very small.

\paragraph{Causes of errors.}

\begin{table}[tbp]
  \centering
  \small
  \setlength{\tabcolsep}{5pt}
  \renewcommand{\arraystretch}{1.1}
  \begin{tabular}{llrrrr}
    \toprule
    Model & System & Total & Incorrect & In SLDNF$\downarrow$ & In loop \\
    \midrule
    Claude       & Standard  & 400 & 5  & 3 & 2 \\
    Claude       & Reasoning & 400 & 0  & 0 & 0 \\
    GPT/o4-mini  & Standard  & 400 & 95 & 2 & 1 \\
    GPT/o4-mini  & Reasoning & 400 & 1  & 0 & 1 \\
    \bottomrule
  \end{tabular}
  \caption{Error breakdown under SLDNF evaluation. SLDNF$\downarrow$ stands for evaluating to a different label than the gold label under SLDNF.}
  \label{tab:sldnf_errors}
\end{table}

Motivated by the observation that most reasoning failures, except for GPT-4.1 in the Standard condition, occur on instances with negation in the rules, we further analysed the evaluation set under standard SLDNF semantics. We identified two classes of instances: (i) instances whose answer under SLDNF differs from the \texttt{PARARULE-Plus} gold label, and (ii) instances for which SLDNF evaluation does not terminate because of recursive rule cycles. Table~\ref{tab:sldnf_errors} reports the breakdowns.

For Claude, all five Standard errors fall into one of these two SLDNF-related categories: three correspond to instances where SLDNF gives a different label from the dataset semantics, and two correspond to non-terminating SLDNF evaluations. The single o4-mini Reasoning error also occurs on an instance that loops under SLDNF. By contrast, GPT-4.1 Standard has many errors that are not explained by these categories, which is consistent with the earlier observation that its failures are not limited to the dataset's non-standard negation semantics. These findings suggest that SLDNF-related semantic mismatch may be the dominant residual failure mode for the strongest systems, but the depth-stratified sample contains too few activating instances (6 diverging answers and 15 non-terminating evaluations) to test this directly. We therefore complement the present evaluation with a targeted set of SLDNF-diverging instances in Section~\ref{sec:not-sldnf}.

\paragraph{Case study.}
On instance \texttt{NegationRule-D5-28653}, with query \texttt{Charlie is smart} and gold answer \emph{true}, the two reasoning models diverge. The intended derivation requires treating \texttt{Charlie is dull} as unresolved against the initial facts (it is not stated), so \texttt{Charlie is not dull} evaluates to true, after which two rules (\textit{if someone is not dull then they are kind}; \textit{if someone is kind then they are smart}) yield the conclusion. Claude follows exactly this chain. GPT/o4-mini instead pursues \texttt{Charlie is dull} as an open subgoal, derives it via forward chaining from other rules, and so concludes the query is false---applying standard SLDNF rather than the prompt's initial-facts reading and constituting a reasoning failure.

\paragraph{Discussion.}
Across depths and rule types, explicit reasoning and symbolic formalization both substantially improve over standard LLMs, with the largest gains where the base model is weakest: the GPT-4.1 Standard baseline reaches only 0.762, while o4-mini Reasoning and the GPT-4.1 Formalizer reach 0.998 and 1.000 respectively. Most residual errors (except for GPT-4.1 Standard) occur on negation instances and concentrate on cases that either loop or receive a different label under SLDNF, consistent with the case-study failure mode of defaulting to standard negation-as-failure rather than the dataset's initial-facts reading. The depth-stratified sample contains too few such cases to test this directly, motivating the targeted evaluation that follows.

\subsection{Targeted evaluation: SLDNF-diverging instances}
\label{sec:not-sldnf}

\paragraph{Dataset.}
To test the hypothesis that the residual errors observed in
Section~\ref{sec:depth-stratified} stem from semantic mismatch rather than
reasoning depth, we constructed a dedicated set of SLDNF-diverging instances
from the negation-type portion of \texttt{PARARULE-Plus}. Each instance was
converted to SWI-Prolog and evaluated twice: once with the intended
\texttt{PARARULE-Plus} encoding, in which body-level negation is checked only against the initial fact base,
and once with standard SLDNF
variant. Instances for which the two evaluations produced different Boolean
answers or would loop under SLDNF were labelled \emph{diverging}. We collected 100 such instances at
depth~4 and 100 at depth~5, giving a stratified targeted set of 200 instances.
All instances in this subset have expected answer \emph{false} under the
\texttt{PARARULE-Plus} semantics.

\begin{table}[tbp]
  \centering
  \small
  \caption{Results on the 200 SLDNF-diverging instances at reasoning
           depths 4 and~5.
           All instances have expected answer \emph{false}.
           Acc.\ = accuracy; Tok/q = mean total tokens per question;
           s/q = mean wall-clock seconds per question.}
  \label{tab:not-sldnf}
  \begin{tabular}{lS[table-format=1.2]S[table-format=1.2]S[table-format=1.2]S[table-format=4.1]S[table-format=2.1]}
    \toprule
     System
      & {Acc.\ (all)}
      & {Acc.\ D4}
      & {Acc.\ D5}
      & {Tok/q}
      & {s/q} \\
    \midrule
          \multicolumn{6}{c}{Claude} \\
          \midrule
      Reasoning        & 0.95 & 0.91 & 0.99 & 1345.4 &  9.5 \\    
      Formalizer       & 1.00 & 1.00 & 1.00 & 1548.4 &  6.2 \\
       \midrule
          \multicolumn{6}{c}{GPT-4/o4-mini} \\
          \midrule
      Reasoning      & 0.94 & 0.94 & 0.94 & 2070.0 & 11.9 \\    
      Formalizer      & 0.99 & 0.98 & 1.00 & 1191.7 &  3.1 \\
    \bottomrule
  \end{tabular}
\end{table}

\paragraph{Discussion.}
Table~\ref{tab:not-sldnf} shows that the targeted SLDNF-diverging subset is substantially more challenging for natural-language reasoning than the depth-stratified sample, while the Formalizer remains near-perfect. Claude Reasoning drops from 1.00 to 0.95 and o4-mini Reasoning from 0.998 to 0.94, while the Formalizer is essentially unchanged (1.00 for Claude, 1.00--0.99 for GPT-4.1). The reliability advantage of symbolic delegation is therefore largest precisely where natural-language reasoning is most fragile.

This robustness is structural: the generated program encodes the \texttt{PARARULE-Plus} semantics explicitly, checking body-level negations against the fixed extensional database rather than against the full derivable closure. The solver is never asked to apply raw SLDNF to those negations, so the main difficulty is shifted from inference to producing a faithful formalization. The depth-wise pattern supports the same interpretation: Claude Reasoning is less accurate at depth~4 than at depth~5, and o4-mini Reasoning is stable across both, so the failures are not explained by reasoning depth but by whether the derivation activates a predicate that is absent from the initial facts yet derivable through rules---the configuration in which \texttt{PARARULE-Plus} and SLDNF disagree.

The cost results reinforce the broader reliability--efficiency trade-off: the Formalizer is faster than the Reasoning condition for both model families (6.2\,s vs.\ 9.5\,s for Claude; 3.1\,s vs.\ 11.9\,s for the OpenAI systems), and the GPT-4.1 Formalizer is simultaneously more accurate and substantially faster than o4-mini Reasoning while using fewer tokens. Externalising the formal inference step to a Prolog solver is most beneficial when the task departs from interpretations the model would default to, rather than simply requiring longer reasoning chains.

\section{Conclusions}
\label{sec:conclusion}

We introduced \textsc{PrologMCP}, an MCP-based interface that lets LLM agents delegate symbolic reasoning to a Prolog solver, separating two capabilities often conflated in LLM evaluations: translating an informal problem into a formal representation, and carrying out the resulting inference. The model is responsible for formalization; the solver performs the deductive step.

We evaluated \textsc{PrologMCP} on \texttt{PARARULE-Plus} along two axes: a depth-stratified sample, and a targeted subset that isolates instances where the dataset's negation semantics diverges from standard SLDNF. The Formalizer reached perfect accuracy on the depth-stratified sample for both model families, with the clearest gains over the weaker GPT-4.1 Standard baseline and lower latency than o4-mini Reasoning. On the targeted SLDNF-diverging subset the Formalizer remained near-perfect (1.00 for Claude, 0.99 for GPT-4.1) while the Reasoning conditions dropped to 0.95 and 0.94---exposing a failure mode where models default to standard negation-as-failure even when instructed otherwise. By encoding the intended semantics in the generated Prolog program, the Formalizer sidesteps this and shifts the burden from inference to translation. The approach does not remove the need for accurate semantic translation, motivating future work on validation of generated logic programs and broader benchmark coverage.

\section{Limitations and Future Work}
The current evaluation relies on a single synthetic benchmark and a closed, frontier-only model set; the cross-system comparison is also not architecturally controlled. \textsc{PrologMCP} itself supports only SWI-Prolog, and its lexical sandbox is not a security boundary. Because all instances formalised on the first attempt, only \texttt{consult\_text} and \texttt{run\_goal} were thoroughly exercised. Immediate next steps include broader and harder benchmarks, weaker and open-source models, fine-tuning small models for the formalization step, additional Prolog dialects, container-level isolation, and a routing layer over domain-specialised MCP servers that dispatches sub-problems to the appropriate solver.


\begin{acknowledgments}
 This work was supported by a Leverhulme Trust International Professorship Grant (LIP-2022-001).
\end{acknowledgments}

\section*{Declaration on Generative AI}
During the preparation of this work, the author(s) used Claude Code in order to: Generate code. Further, the author(s) used Claude in order to: Improve writing style. After using these tool(s)/service(s), the author(s) reviewed and edited the content as needed and take(s) full responsibility for the publication's content.

\bibliography{bibliography}

@article{kowalski1971linear,
  author  = {Robert A. Kowalski and Donald Kuehner},
  title   = {Linear Resolution with Selection Function},
  journal = {Artificial Intelligence},
  volume  = {2},
  number  = {3--4},
  pages   = {227--260},
  year    = {1971},
  publisher = {Elsevier}
}

@article{korner2022fifty,
  title   = {Fifty Years of Prolog and Beyond},
  author  = {K{\"o}rner, Philipp and Leuschel, Michael and Barbosa, Jo{\~a}o and Santos Costa, V{\'i}tor and Dahl, Ver{\'o}nica and Hermenegildo, Manuel V. and Morales, Jos{\'e} F. and Wielemaker, Jan and Diaz, Daniel and Abreu, Salvador and Ciatto, Giovanni},
  journal = {Theory and Practice of Logic Programming},
  year    = {2022},
  publisher = {Cambridge University Press}
}

@incollection{AptBlairWalker1988,
  author    = {Krzysztof R. Apt and Howard A. Blair and Adrian Walker},
  title     = {Towards a Theory of Declarative Knowledge},
  booktitle = {Foundations of Deductive Databases and Logic Programming},
  editor    = {Jack Minker},
  publisher = {Morgan Kaufmann},
  year      = {1988},
  pages     = {89--148}
}

@incollection{clark1978negation,
  author    = {Keith L. Clark},
  title     = {Negation as Failure},
  booktitle = {Logic and Data Bases},
  editor    = {Herv{\'e} Gallaire and Jack Minker},
  pages     = {293--322},
  year      = {1978},
  publisher = {Springer}
}

@inproceedings{mensfelt2026towards,
title = "Towards a Common Framework for Autoformalization",
author = "Agnieszka Mensfelt and Tena Cucala, David and Santiago Franco and Angeliki Koutsoukou-Argyraki and Vince Trencsenyi and Kostas Stathis",
year = "2026",
month = mar,
day = "14",
doi = "10.1609/aaai.v40i48.42132",
language = "English",
volume = "40",
pages = "40971--40980",
booktitle = "Proceedings of the AAAI Conference on Artificial Intelligence",
}

@article{zunjare2026neuroprolog,
  title={NeuroProlog: Multi-Task Fine-Tuning for Neurosymbolic Mathematical Reasoning via the Cocktail Effect},
  author={Zunjare, Pratibha and Hsiao, Michael},
  journal={arXiv preprint arXiv:2603.02504},
  year={2026}
}

@inproceedings{szeider2025bridging,
  title={Bridging language models and symbolic solvers via the model context protocol},
  author={Szeider, Stefan},
  booktitle={28th International Conference on Theory and Applications of Satisfiability Testing (SAT 2025)},
  pages={30--1},
  year={2025},
  organization={Schloss Dagstuhl--Leibniz-Zentrum f{\"u}r Informatik}
}

@article{mellgren2025training,
  title={Training Language Models to Use Prolog as a Tool},
  author={Mellgren, Niklas and Schneider-Kamp, Peter and Poech, Lukas Galke},
  journal={arXiv preprint arXiv:2512.07407},
  year={2025}
}

@inproceedings{lin2025zebralogic,
  title={ZebraLogic: On the Scaling Limits of LLMs for Logical Reasoning},
  author={Lin, Bill Yuchen and Le Bras, Ronan and Richardson, Kyle and Sabharwal, Ashish and Poovendran, Radha and Clark, Peter and Choi, Yejin},
  booktitle={International Conference on Machine Learning},
  pages={37889--37905},
  year={2025},
  organization={PMLR}
}

@inproceedings{mirzadeh2025gsm,
title={{GSM}-Symbolic: Understanding the Limitations of Mathematical Reasoning in Large Language Models},
author={Seyed Iman Mirzadeh and Keivan Alizadeh and Hooman Shahrokhi and Oncel Tuzel and Samy Bengio and Mehrdad Farajtabar},
booktitle={The Thirteenth International Conference on Learning Representations},
year={2025},
url={https://openreview.net/forum?id=AjXkRZIvjB}
}

@inproceedings{kazemi2025bbeh,
  title={Big-bench extra hard},
  author={Kazemi, Mehran and Fatemi, Bahare and Bansal, Hritik and Palowitch, John and Anastasiou, Chrysovalantis and Mehta, Sanket Vaibhav and Jain, Lalit K and Aglietti, Virginia and Jindal, Disha and Chen, Yuanzhu Peter and others},
  booktitle={Proceedings of the 63rd Annual Meeting of the Association for Computational Linguistics (Volume 1: Long Papers)},
  pages={26473--26501},
  year={2025}
}

@article{dziri2023faith,
  title={Faith and fate: Limits of transformers on compositionality},
  author={Dziri, Nouha and Lu, Ximing and Sclar, Melanie and Li, Xiang Lorraine and Jiang, Liwei and Lin, Bill Yuchen and Welleck, Sean and West, Peter and Bhagavatula, Chandra and Le Bras, Ronan and others},
  journal={Advances in neural information processing systems},
  volume={36},
  pages={70293--70332},
  year={2023}
}

@article{shojaee2025illusion,
  title={The illusion of thinking: Understanding the strengths and limitations of reasoning models via the lens of problem complexity},
  author={Shojaee, Parshin and Mirzadeh, Iman and Horton, Maxwell and Bengio, Samy and Farajtabar, Mehrdad and others},
  journal={Advances in Neural Information Processing Systems},
  volume={38},
  pages={108018--108059},
  year={2026}
}

@article{schick2023toolformer,
  title={Toolformer: Language models can teach themselves to use tools},
  author={Schick, Timo and Dwivedi-Yu, Jane and Dess{\`\i}, Roberto and Raileanu, Roberta and Lomeli, Maria and Hambro, Eric and Zettlemoyer, Luke and Cancedda, Nicola and Scialom, Thomas},
  journal={Advances in neural information processing systems},
  volume={36},
  pages={68539--68551},
  year={2023}
}

@inproceedings{gao2023pal,
  title={Pal: Program-aided language models},
  author={Gao, Luyu and Madaan, Aman and Zhou, Shuyan and Alon, Uri and Liu, Pengfei and Yang, Yiming and Callan, Jamie and Neubig, Graham},
  booktitle={International conference on machine learning},
  pages={10764--10799},
  year={2023},
  organization={PMLR}
}

@article{borazjanizadeh2024reliable,
  title={Reliable reasoning beyond natural language},
  author={Borazjanizadeh, Nasim and Piantadosi, Steven T},
  journal={arXiv preprint arXiv:2407.11373},
  year={2024}
}

@inproceedings{yang2024arithmetic,
  title={Arithmetic reasoning with LLM: Prolog generation \& permutation},
  author={Yang, Xiaocheng and Chen, Bingsen and Tam, Yik-Cheung},
  booktitle={Proceedings of the 2024 Conference of the North American Chapter of the Association for Computational Linguistics: Human Language Technologies (Volume 2: Short Papers)},
  pages={699--710},
  year={2024}
}

@article{tan2024thought,
  title={Thought-like-pro: Enhancing reasoning of large language models through self-driven prolog-based chain-of-thought},
  author={Tan, Xiaoyu and Deng, Yongxin and Qiu, Xihe and Xu, Weidi and Qu, Chao and Chu, Wei and Xu, Yinghui and Qi, Yuan},
  journal={arXiv preprint arXiv:2407.14562},
  year={2024}
}

@article{he2025protoreasoning,
  title={Protoreasoning: Prototypes as the foundation for generalizable reasoning in llms},
  author={He, Feng and Chen, Zijun and Liang, Xinnian and Ma, Tingting and Qiu, Yunqi and Wu, Shuangzhi and Yan, Junchi},
  journal={arXiv preprint arXiv:2506.15211},
  year={2025}
}

@MISC{anthropic2024mcp,
  title        = "{Introducing the Model Context Protocol}",
  author       = "{Anthropic}",
  year         = {2024},
  howpublished = "\url{https://www.anthropic.com/news/model-context-protocol}"
}

@ARTICLE{Hou2026e,
  title     = "{Model Context Protocol (MCP): Landscape, security threats, and
               future research directions}",
  author    = "Hou, Xinyi and Zhao, Yanjie and Wang, Shenao and Wang, Haoyu",
  journal   = "ACM Transactions on Software Engineering and Methodology",
  publisher = "Association for Computing Machinery (ACM)",
  number    =  3796519,
  year      =  2026
}

@article{bao2022multi,
  title={Multi-step deductive reasoning over natural language: An empirical study on out-of-distribution generalisation},
  author={Bao, Qiming and Peng, Alex Yuxuan and Hartill, Tim and Tan, Neset and Deng, Zhenyun and Witbrock, Michael and Liu, Jiamou},
  journal={arXiv preprint arXiv:2207.14000},
  year={2022}
}

@article{wielemaker2012swi,
  title={Swi-prolog},
  author={Wielemaker, Jan and Schrijvers, Tom and Triska, Markus and Lager, Torbj{\"o}rn},
  journal={Theory and Practice of Logic Programming},
  volume={12},
  number={1-2},
  pages={67--96},
  year={2012},
  publisher={Cambridge University Press}
}

@article{wu_autoformalization_2022,
	title = {Autoformalization with {Large} {Language} {Models}},
	volume = {35},
	url = {https://proceedings.neurips.cc/paper_files/paper/2022/hash/d0c6bc641a56bebee9d985b937307367-Abstract-Conference.html},
	language = {en},
	urldate = {2025-03-23},
	journal = {Advances in Neural Information Processing Systems},
	author = {Wu, Yuhuai and Jiang, Albert Qiaochu and Li, Wenda and Rabe, Markus and Staats, Charles and Jamnik, Mateja and Szegedy, Christian},
	month = dec,
	year = {2022},
	pages = {32353--32368},
}

@book{abiteboul1995foundations,
  title={Foundations of databases},
  author={Abiteboul, Serge and Hull, Richard and Vianu, Victor},
  volume={8},
  year={1995},
  publisher={Addison-Wesley Reading}
}

@techreport{mccarthy1959,
author = {McCarthy, John},
title = {Programs with Common Sense},
year = {1960},
publisher = {Massachusetts Institute of Technology},
address = {USA}
}

@article{newell1976, 
    author={Allen Newell and Herbert A. Simon}, 
    title={Computer science as empirical inquiry: Symbols and search}, 
    journal={Communications of the ACM}, 
    volume={19}, 
    number={3}, 
    pages={113--126}, 
    year={1976}
}

@inproceedings{kowalski1974, 
    author={Robert Kowalski}, 
    title={Predicate logic as a programming language}, 
    booktitle={IFIP Congress}, 
    pages={569--574}, 
    year={1974}
}

@INPROCEEDINGS{Clark2020p,
  title     = "{Transformers as soft reasoners over language}",
  author    = "Clark, Peter and Tafjord, Oyvind and Richardson, Kyle",
  booktitle = "{Proceedings of the Twenty-Ninth International Joint Conference on Artificial Intelligence (IJCAI-PRICAI 2020)}",
  publisher = "International Joint Conferences on Artificial Intelligence Organization",
  address   = "California",
  year      =  2020
}

@inproceedings{mensfelt2024generative,
    author={Agnieszka Mensfelt and Kostas Stathis and Vince Tencsenyi},
    title={Generative Agents for Multi-Agent Autoformalization of Interaction Scenarios},
    year={2025},
  booktitle = {Proceedings of the 28th European Conference on Artificial Intelligence (ECAI 2025)},
  series    = {Frontiers in Artificial Intelligence and Applications},
  volume    = {413},
  pages     = {3759--3766},
  publisher = {IOS Press},
  year      = {2025},
  doi       = {10.3233/FAIA251256},
}

@misc{rybinski,
  author       = {{adamrybinski}},
  title        = {Prolog-{MCP} {S}erver},
  howpublished = {\url{https://github.com/adamrybinski/prolog-mcp}},
  year         = {2025},
  note         = {GitHub repository. Accessed 2026-06-19}
}

@misc{umuro,
  author       = {{umuro}},
  title        = {prolog-mcp},
  howpublished = {\url{https://github.com/umuro/prolog-mcp}},
  year         = {2025},
  note         = {GitHub repository. Accessed 2026-06-19}
}

@misc{vpursuit,
  author       = {{vpursuit}},
  title        = {{MCP} {E}cosystem},
  howpublished = {\url{https://github.com/vpursuit/model-context-lab}},
  year         = {2025},
  note         = {GitHub repository. Accessed 2026-06-19}
}

@misc{rikarazome,
  author       = {{rikarazome}},
  title        = {prolog-reasoner},
  howpublished = {\url{https://github.com/rikarazome/prolog-reasoner}},
  year         = {2026},
  note         = {GitHub repository and {PyPI} package. Accessed 2026-06-19}
}

\newpage

\appendix
\definecolor{toolHeader}{HTML}{1F3A5F}
\definecolor{toolBody}{HTML}{F5F7FB}
\definecolor{toolAccent}{HTML}{2E5C8A}
\definecolor{schemaBg}{HTML}{EEF2F7}
\definecolor{respBg}{HTML}{FBF7EE}
\definecolor{respAccent}{HTML}{8A6A2E}
\definecolor{promptHeader}{HTML}{2E5C3F}
\definecolor{promptBody}{HTML}{F4F9F5}

\newtcolorbox{toolbox}[2]{%
  enhanced, breakable,
  colback=toolBody, colframe=toolHeader,
  colbacktitle=toolHeader, coltitle=white,
  fonttitle=\bfseries\ttfamily,
  title={#1\hfill\normalfont\itshape\small\color{white!85!toolHeader} #2},
  boxrule=0.6pt, arc=2pt,
  left=8pt, right=8pt, top=6pt, bottom=6pt,
  toptitle=3pt, bottomtitle=3pt,
  before skip=10pt, after skip=10pt,
}

\newtcolorbox{schemabox}{%
  enhanced, breakable,
  colback=schemaBg, colframe=schemaBg,
  boxrule=0pt, arc=1.5pt,
  left=6pt, right=6pt, top=4pt, bottom=4pt,
  before skip=4pt, after skip=4pt,
}

\newtcolorbox{respbox}{%
  enhanced, breakable,
  colback=respBg, colframe=respAccent,
  boxrule=0pt, leftrule=2.5pt,
  arc=1pt,
  left=8pt, right=6pt, top=4pt, bottom=4pt,
  before skip=4pt, after skip=4pt,
}

\newtcolorbox{promptbox}[1]{%
  enhanced, breakable,
  colback=promptBody, colframe=promptHeader,
  colbacktitle=promptHeader, coltitle=white,
  fonttitle=\bfseries,
  title={#1},
  boxrule=0.6pt, arc=2pt,
  left=8pt, right=8pt, top=6pt, bottom=6pt,
  toptitle=3pt, bottomtitle=3pt,
  before skip=10pt, after skip=10pt,
}

\newcommand{\rkeys}{\textbf{\textcolor{respAccent}{Response keys.}}\ }

\section{PARARULE-Plus}
\label{app:examples}

\vspace{1em}

\begin{tcolorbox}[
  title={Example instance: non-negation, people},
  colback=gray!3,
  colframe=black!70,
  fonttitle=\bfseries,
  breakable,
  sharp corners,
  boxrule=0.6pt
]
\textbf{Instance ID:} \texttt{NonNegationRule-D2-11112}

\medskip
\textbf{Context.}
Charlie is strong. Charlie is high. Charlie is huge. Bob is thin. Bob is small. Erin is quiet. Erin is smart. Erin is kind. Anne is bad. Anne is sad. Anne is rough. Strong people are quiet. If someone is thin and small then they are short. If someone is bad and sad then they are poor. If someone is quiet and smart then they are wealthy. All short people are little. All quiet people are smart. All wealthy people are nice. All poor people are dull.
\medskip
\textbf{Query.}
Charlie is not smart.
\end{tcolorbox}

\vspace{1em}

\begin{tcolorbox}[
  title={Example instance: negation, people},
  colback=gray!3,
  colframe=black!70,
  fonttitle=\bfseries,
  breakable,
  sharp corners,
  boxrule=0.6pt
]
\textbf{Instance ID:} \texttt{NegationRule-D2-6305}

\medskip
\textbf{Context.}
Alan is big. Alan is high. Gary is small. Gary is thin. Fiona is smart. Harry is bad. Harry is poor. If someone is not strong then they are bad. If someone is not sad then they are nice. If someone is smart then they are kind. If someone is kind and not rough then they are quiet. If someone is bad and not strong then they are dull. If someone is small and thin then they are rough. If someone is rough and not kind then they are poor. All nice people are wealthy.
\medskip
\textbf{Query.}
Fiona is quiet.
\end{tcolorbox}

\vspace{1em}

\begin{tcolorbox}[
  title={Example instance: non-negation, animal},
  colback=gray!3,
  colframe=black!70,
  fonttitle=\bfseries,
  breakable,
  sharp corners,
  boxrule=0.6pt
]
\textbf{Instance ID:} \texttt{NonNegationRule-Animal-D2-13824}

\medskip
\textbf{Context.}
The wolf is dull. The wolf is sleepy. The wolf is slow. The wolf sees the mouse. The bald eagle chases the rabbit. The bald eagle is heavy. The bald eagle is big. The mouse is smart. The mouse is kind. The mouse is round. The rabbit is lovely. The rabbit is small. The rabbit is cute. Smart animals are lovely. If something is sleepy then it attacks the mouse. If something attacks the mouse then it is rough. If something is dull and sleepy then it is slow. If something is lovely and small then it is furry. If something is heavy and big then it is awful. All slow animals are lazy. All lovely animals are small. All awful animals are strong. All furry animals are beautiful.

\medskip
\textbf{Query.}
The wolf is not lazy.
\end{tcolorbox}

\vspace{1em}

\begin{tcolorbox}[
  title={Example instance: negation, animal},
  colback=gray!3,
  colframe=black!70,
  fonttitle=\bfseries,
  breakable,
  sharp corners,
  boxrule=0.6pt
]
\textbf{Instance ID:} \texttt{NegationRule-Animal-D2-3081}

\medskip
\textbf{Context.}
The bald eagle is sleepy. The bald eagle is rough. The leopard is heavy.
The leopard is fierce. The bald eagle visits the rabbit. The leopard sees the dog.
The rabbit is nice. The dog is nice. The dog is furry. The dog is lovely.
If something is not nice then it needs the rabbit. If something needs the rabbit then it is slow.
If something is not round then it is heavy. If something is not strong then it is cute.
If something is furry then it is lovely. If something is lovely and not big then it is small.
If something is heavy and not round then it is awful. If something is sleepy and rough then it is big.
If something is big and not lovely then it is fierce. All cute animals are beautiful.

\medskip
\textbf{Query.}
The bald eagle is awful.
\end{tcolorbox}

\newpage

\section{Tool schemas}
\label{app:tools}

\begin{toolbox}{consult\_text}{load Prolog source into a new session}
Returns a \texttt{load\_id} used by all other tools.

\begin{schemabox}
\begin{verbatim}
{
  "type": "object",
  "properties": {
    "source_text":         {"type": "string"},
    "dialect":             {"type": "string", "enum": ["swi"], "default": "swi"},
    "module_name":         {"type": "string"},
    "include_predicates":  {"type": "boolean", "default": false}
  },
  "required": ["source_text"]
}
\end{verbatim}
\end{schemabox}

\begin{respbox}
\rkeys
\texttt{load\_id} (string),
\texttt{module\_name} (string),
\texttt{dialect} (\texttt{"swi"}),
\texttt{messages} (list of \texttt{\{severity, kind, message, line, column\}}),
and --- when \texttt{include\_predicates} is \texttt{true} ---
\texttt{predicates} (list of predicate descriptors).
Compile-time errors are reported through \texttt{messages} rather
than as a top-level error: the session is created so the agent can
inspect and repair.
\end{respbox}
\end{toolbox}

\begin{toolbox}{run\_goal}{execute a goal in an existing session}
\begin{schemabox}
\begin{verbatim}
{
  "type": "object",
  "properties": {
    "load_id":            {"type": "string"},
    "goal":               {"type": "string"},
    "max_answers":        {"type": "integer", "default": 10},
    "time_limit":         {"type": "number",  "default": 30.0},
    "max_depth":          {"type": "integer", "default": 1000000},
    "allow_side_effects": {"type": "boolean", "default": true}
  },
  "required": ["load_id", "goal"]
}
\end{verbatim}
\end{schemabox}

\begin{respbox}
\rkeys
\texttt{outcome} (\texttt{"success"} \(\mid\) \texttt{"failure"} \(\mid\)
\texttt{"timeout"} \(\mid\) \texttt{"depth\_exceeded"}),
\texttt{answers} (list of variable-binding maps with typed values),
\texttt{output} (any captured \texttt{stdout} from the goal),
\texttt{truncated} (boolean: \texttt{true} when more answers were
available than \texttt{max\_answers} returned).
Goal exceptions are returned as a top-level error with
\texttt{kind} drawn from
\texttt{type\_error}, \texttt{existence\_error}, \texttt{instantiation\_error},
\texttt{permission\_error}, or \texttt{execution\_exception}.
\end{respbox}
\end{toolbox}

\begin{toolbox}{inspect\_predicate}{return the live definition of a named predicate}
\begin{schemabox}
\begin{verbatim}
{
  "type": "object",
  "properties": {
    "load_id": {"type": "string"},
    "name":    {"type": "string"},
    "arity":   {"type": "integer"}
  },
  "required": ["load_id", "name", "arity"]
}
\end{verbatim}
\end{schemabox}

\begin{respbox}
\rkeys
\texttt{found} (boolean),
\texttt{type} (\texttt{static} \(\mid\) \texttt{dynamic} \(\mid\)
\texttt{foreign} \(\mid\) \texttt{undefined}),
\texttt{exported} (boolean),
\texttt{clause\_count} (integer),
\texttt{clauses} (list of source-form clauses).
On miss, \texttt{candidates} lists predicates whose name matches up
to fuzzy comparison, suitable for repair.
\end{respbox}
\end{toolbox}

\begin{toolbox}{list\_messages}{return diagnostics accumulated for a session}
\begin{schemabox}
\begin{verbatim}
{
  "type": "object",
  "properties": {
    "load_id":      {"type": "string"},
    "min_severity": {"type": "string", "enum": ["info", "warning", "error"],
                                       "default": "info"}
  },
  "required": ["load_id"]
}
\end{verbatim}
\end{schemabox}

\begin{respbox}
\rkeys
\texttt{messages} (list of \texttt{\{severity, kind, message, line, column\}}
records). Messages accumulate across all calls in the session; the
caller filters by \texttt{min\_severity}.
\end{respbox}
\end{toolbox}

\begin{toolbox}{run\_tests}{run \texttt{plunit} tests in the session}
Optionally loads additional test source first.

\begin{schemabox}
\begin{verbatim}
{
  "type": "object",
  "properties": {
    "load_id":          {"type": "string"},
    "test_source_text": {"type": "string"}
  },
  "required": ["load_id"]
}
\end{verbatim}
\end{schemabox}

\begin{respbox}
\rkeys
\texttt{passed}, \texttt{failed}, \texttt{skipped}, \texttt{total}
(integers);
\texttt{results} (list of per-test records with name, status, and
failure message if applicable).
\end{respbox}
\end{toolbox}

\begin{toolbox}{trace\_goal}{trace a goal via a depth-bounded meta-interpreter}
Returns a structured proof tree.

\begin{schemabox}
\begin{verbatim}
{
  "type": "object",
  "properties": {
    "load_id":   {"type": "string"},
    "goal":      {"type": "string"},
    "max_depth": {"type": "integer", "default": 10},
    "max_nodes": {"type": "integer", "default": 200}
  },
  "required": ["load_id", "goal"]
}
\end{verbatim}
\end{schemabox}

\begin{respbox}
\rkeys
\texttt{tree} (nested object with \texttt{goal}, \texttt{rule},
\texttt{children}),
\texttt{node\_count} (integer).
Implemented through a custom meta-interpreter rather than
SWI-Prolog's tracer; output is structured rather than line-oriented.
\end{respbox}
\end{toolbox}

\begin{toolbox}{replace\_predicate}{replace the clauses of a single predicate}
\texttt{source\_text} must define only the target name/arity.

\begin{schemabox}
\begin{verbatim}
{
  "type": "object",
  "properties": {
    "load_id":     {"type": "string"},
    "name":        {"type": "string"},
    "arity":       {"type": "integer"},
    "source_text": {"type": "string"}
  },
  "required": ["load_id", "name", "arity", "source_text"]
}
\end{verbatim}
\end{schemabox}

\begin{respbox}
\rkeys
\texttt{predicate} (echo of \texttt{name}/\texttt{arity}),
\texttt{messages} (list of diagnostics from the patch load, with
line and column locators referring to the replacement source).
The session is left in its prior state if the patch fails to
compile; partial replacement is never observed.
\end{respbox}
\end{toolbox}

\begin{toolbox}{get\_source}{return the current source of a session}
Reconstructed from all live clauses; reflects any replacements made
since load.

\begin{schemabox}
\begin{verbatim}
{
  "type": "object",
  "properties": {
    "load_id": {"type": "string"}
  },
  "required": ["load_id"]
}
\end{verbatim}
\end{schemabox}

\begin{respbox}
\rkeys
\texttt{source} (string).
\end{respbox}
\end{toolbox}

\begin{toolbox}{close\_session}{terminate a session and release resources}
Releases the subprocess and the session's temporary directory.

\begin{schemabox}
\begin{verbatim}
{
  "type": "object",
  "properties": {
    "load_id": {"type": "string"}
  },
  "required": ["load_id"]
}
\end{verbatim}
\end{schemabox}

\begin{respbox}
\rkeys
\texttt{closed} (boolean).
Sessions are also released automatically when the registry's TTL
expires or when LRU eviction triggers; explicit close is a courtesy
for long-running agents.
\end{respbox}
\end{toolbox}

\newpage
\section{Experimental prompts}
\label{app:prompts}

\subsection{Formalizer-agent system prompt}
\label{app:prompts:formalize}

\begin{promptbox}{Formalizer agent}
\begin{verbatim}
You are a mechanical translator from English to SWI-Prolog.  Your job is \
purely syntactic: map each English sentence to its Prolog equivalent \
one-by-one.  Do not trace, evaluate, or solve the reasoning problem.

Given a CONTEXT (facts and rules) and a yes/no QUESTION, produce:
  1. Valid SWI-Prolog source code that captures every fact and rule.
  2. A single query goal:
       - goal_true  -- SUCCEEDS when the answer is TRUE
       If goal_true fails, the answer is FALSE (closed-world assumption:
       anything not provable from the given facts and rules is false).

SWI-Prolog conventions:
  - All atom names are lowercase and use underscores for spaces:
      `anne`, `is_kind`, `likes_fish`
  - Facts:  `kind(anne).`   `likes(bob, alice).`
  - Rules:  `nice(X) :- kind(X).`
  - Negation-as-failure: `\\+`  (e.g. `not_big(X) :- \\+ big(X).`)
  - Variables start with uppercase: `X`, `Y`, `Who`
  - Do NOT add a module declaration; the server handles namespacing.
  - Do NOT add ``:- discontiguous`` declarations; the server injects them
    automatically for any predicate with multiple clauses.
  - For every predicate that appears ONLY inside ``\\+`` and is not defined
    by any fact or rule, add ``:- dynamic pred/arity.`` -- otherwise
    SWI-Prolog throws existence_error instead of failing cleanly.

Negation -- critical rule (no exceptions, apart from goal):
  Every ``\\+`` in a rule body MUST check ``initially/2``, never the derived
  predicate directly.  Do not reason about whether a predicate could be
  derived -- apply this rule unconditionally to every negation:
  - Assert every base fact in TWO forms: ``pred(entity).`` and
    ``initially(pred, entity).``
  - Write ``\\+ initially(pred, X)`` for EVERY negation in every rule body.

  Example -- "if someone is not high then they are dull":
    dull(X) :- \\+ initially(high, X).   % correct
    dull(X) :- \\+ high(X).              % WRONG -- high/1 may be derived

  Example -- "if someone is sad and not quiet then they are rough":
    rough(X) :- sad(X), \\+ initially(quiet, X).   % correct
    rough(X) :- sad(X), \\+ quiet(X).              % WRONG -- quiet/1 may be derived

Do **not** put ``initially/2`` in the goal.
   Example goal -- "Anne is not sad."
     \\+sad(anne).


Response format -- output ONLY the XML tags below, no reasoning text:

<prolog>
% facts and rules
fact(entity).  initially(fact, entity).
conclusion(X) :- premise(X).
</prolog>
<goal_true>query_predicate(entity)</goal_true>
\end{verbatim}
\end{promptbox}

\subsection{Standard-LLM (and Reasoning) system prompt}
\label{app:prompts:standard}

\begin{promptbox}{Standard LLM \textnormal{\&} reasoning}
\begin{verbatim}
You are a logical reasoning expert.  Given a context (a set of facts and \
rules) and a yes/no question, determine the answer using only the information \
provided.  Use the closed-world assumption: if something cannot be derived \
from the given facts and rules, treat it as false.

Negation has two different readings depending on where it appears.

(1) Negation inside a rule body -- initial-facts reading.
When a rule condition says "not X", check only whether X appears as a \
directly stated fact in the initial context -- not whether X can be derived \
through rules.  Even if a person acquires property X through a chain of \
rules, they still satisfy "not X" in other rule bodies, because X was not \
initially stated for them.
  Example: if "rough" is not stated for Anne but can later be derived, then \
the rule "if someone is not rough then they are smart" still fires for \
Anne -- she counts as "not rough" because rough was not an initial fact.

(2) Negation in the question -- full-derivation reading.
When the question itself asks whether "not X" holds of someone, use \
both facts and rules: "not X(a)" is true \
iff X(a) cannot be derived from the facts and rules (whether directly \
stated or obtained through any chain of rule applications).  Do not apply \
the initial-facts reading here -- it is specific to rule bodies.
  Example: if "energetic" is not stated for Bob but can be derived for \
Bob via some rule chain, then the question "is Bob not energetic?" is \
false, because energetic(Bob) is derivable.  The question "is Bob not \
energetic?" is true only if energetic(Bob) is neither stated nor \
derivable.

Procedure: when evaluating rule bodies, freeze the set of initial facts \
and check negated conditions against that frozen set; when evaluating the \
question, run the full derivation and check negated questions against the \
closure.

Reply with exactly one word: true or false.
  - true   -- the question follows from the context
  - false  -- the question does not follow from the context
\end{verbatim}
\end{promptbox}

\end{document}